\documentclass[lettersize,journal]{IEEEtran}
\usepackage{graphicx}
\usepackage{cite}
\usepackage{amsmath, amssymb, amsfonts}
\usepackage{array}
\usepackage{bm}
\usepackage{booktabs}
\usepackage{ulem}
\usepackage{comment}
\usepackage{verbatim}
\usepackage{url}
\usepackage{stfloats}
\usepackage[linesnumbered, ruled, vlined]{algorithm2e}
\usepackage{caption}
\usepackage{subcaption}
\usepackage{hyperref}
\usepackage{cleveref}
\usepackage{arydshln}

\begin{document}

\title{Overcoming Negative Transfer by Online Selection: Distant Domain Adaptation for Fault Diagnosis}

\author{Ziyan Wang, Mohamed Ragab, Wenmian Yang, Min Wu~\IEEEmembership{Senior Member,~IEEE,}, Sinno Jialin Pan, Jie Zhang, Zhenghua Chen~\IEEEmembership{Senior Member,~IEEE,}
     
\thanks{
Ziyan Wang and Jie Zhang are with the School of Computer Science and Engineering, Nanyang Technological University,
Singapore 639798, (E-mail: wang1753@e.ntu.edu.sg; zhangj@ntu.edu.sg).

Mohamed Ragab, Min Wu, and Zhenghua Chen are with the Institute for Infocomm
Research, Agency for Science, Technology and Research, Singapore 138632, (E-mail: mohamedr002@e.ntu.edu.sg; wumin@i2r.a-star.edu.sg; chen0832@e.ntu.edu.sg). (Corresponding author:
Zhenghua Chen.)

Wenmian Yang is with the Advanced Institute of Natural Sciences, Beijing Normal University, China 519087, (E-mail: wenmianyang@bnu.edg.cn).

Sinno Jialin Pan is with the Department of Computer Science and Engineering, The Chinese University of Hong Kong, China 999077, (E-mail: sinnopan@cse.cuhk.edu.hk)
}
}

\markboth{Journal of \LaTeX\ Class Files,~Vol.~14, No.~8, August~2021}%
{Shell \MakeLowercase{\textit{et al.}}: A Sample Article Using IEEEtran.cls for IEEE Journals}


\maketitle

\begin{abstract}
Unsupervised domain adaptation (UDA) has achieved remarkable success in fault diagnosis, bringing significant benefits to diverse industrial applications.  While most UDA methods focus on cross-working condition scenarios where the source and target domains are notably similar, real-world applications often grapple with severe domain shifts. We coin the term `distant domain adaptation problem' to describe the challenge of adapting from a labeled source domain to a significantly disparate unlabeled target domain. This problem exhibits the risk of negative transfer, where extraneous knowledge from the source domain adversely affects the target domain performance. Unfortunately, conventional UDA methods often falter in mitigating this negative transfer, leading to suboptimal performance. In response to this challenge, we propose a novel Online Selective Adversarial Alignment (OSAA) approach. Central to OSAA is its ability to dynamically identify and exclude distant source samples via an online gradient masking approach, focusing primarily on source samples that closely resemble the target samples. Furthermore, recognizing the inherent complexities in bridging the source and target domains, we construct an intermediate domain to act as a transitional domain and ease the adaptation process. Lastly, we develop a class-conditional adversarial adaptation to address the label distribution disparities while learning domain invariant representation to account for potential label distribution disparities between the domains. Through detailed experiments and ablation studies on two real-world datasets, we validate the superior performance of the OSAA method over state-of-the-art methods, underscoring its significant utility in practical scenarios with severe domain shifts. 
\end{abstract}

\begin{IEEEkeywords}
Distant domain adaptation, fault diagnosis, deep learning, time series, rotating machines.
\end{IEEEkeywords}

\section{Introduction}
\label{sec:introduction}
\IEEEPARstart{I}{ntelligent} fault diagnosis is crucial in maintaining modern machinery, optimizing operational efficiency, minimizing downtime, and safeguarding standards in various settings. 
Deep learning has achieved notable success in identifying machine faults \cite{b.2, b.3, b.4}, but this largely relies on the availability of extensive labeled data, often scarce in practical scenarios \cite{b.5}. 
To combat this, Unsupervised Domain Adaptation (UDA) has emerged, utilizing knowledge from labeled source domains to enhance performance in unlabeled, shifted target domains \cite{b.6}. 

\begin{figure}[h]
    \centering
    \includegraphics[width=0.43\textwidth]{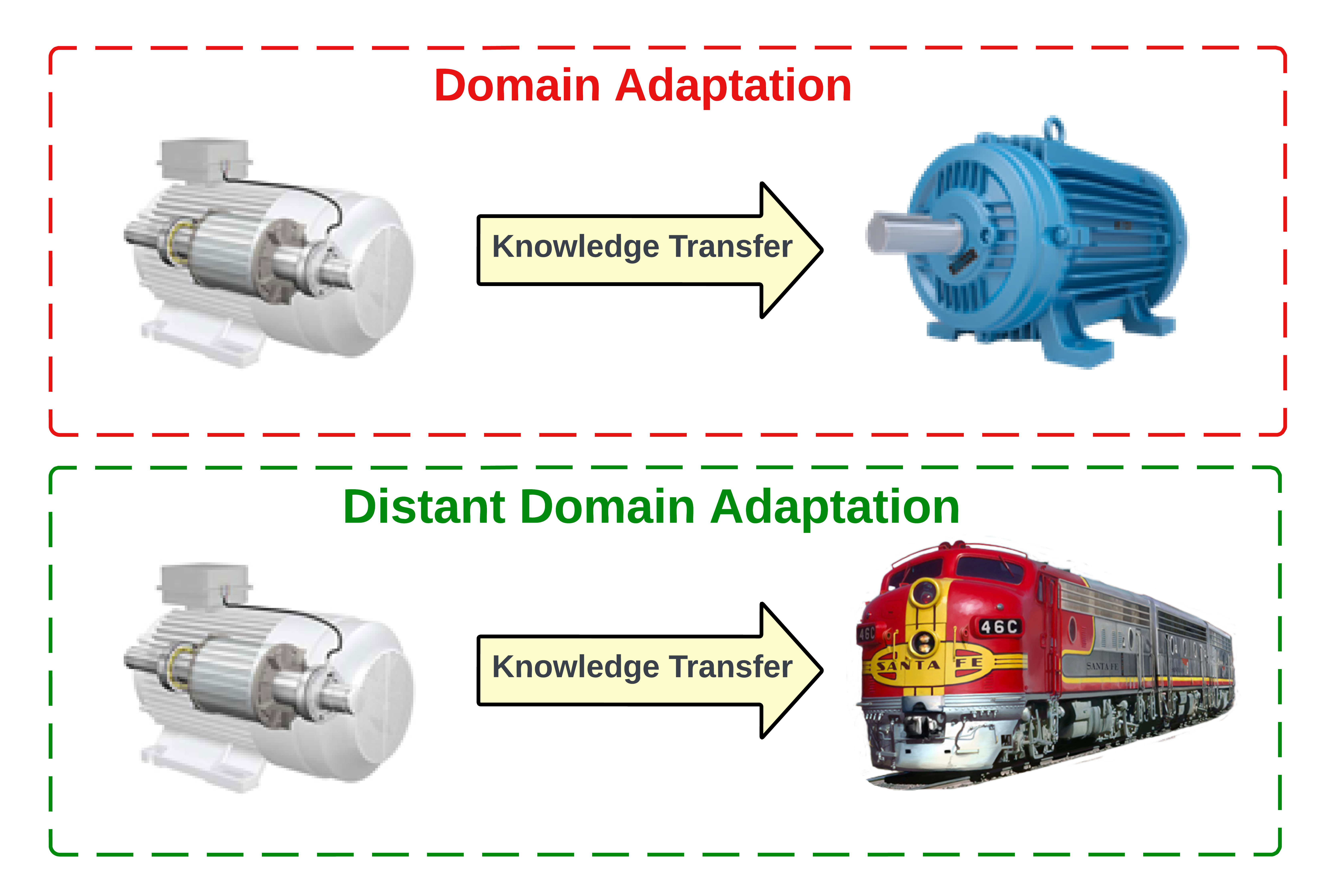}
    \caption{Classic domain shift versus distant domain shift.}
    \label{fig:enter-label}
\end{figure}
Several methodologies have effectively implemented UDA in fault diagnosis \cite{b.8, b.9, b.new}, primarily focusing on machines with similar specifications under varying working conditions, assuming considerable similarity between domains.  
However, in practical applications, it is not uncommon to confront scenarios where substantial discrepancies exist between the source and target domains. Consider, for example, the case depicted in Fig. \ref{fig:enter-label}. Faults in lab machines, operating under predefined configurations, differ starkly from those in industrial machines subjected to complex and noisy environments.
With such severe domain shifts, traditional domain adaptation methods often falter, sometimes yielding even inferior results compared to non-transfer methods. We argue that the downfall of conventional UDA methods is mainly attributed to the negative impact of features learned from distant source samples on the target domain adaptation, a scenario we term as a `distant domain adaptation' problem. Despite its prevalence in practical situations, this issue remains relatively less explored. 

\begin{figure}[h]
    \centering
    \includegraphics[width=0.49\textwidth]{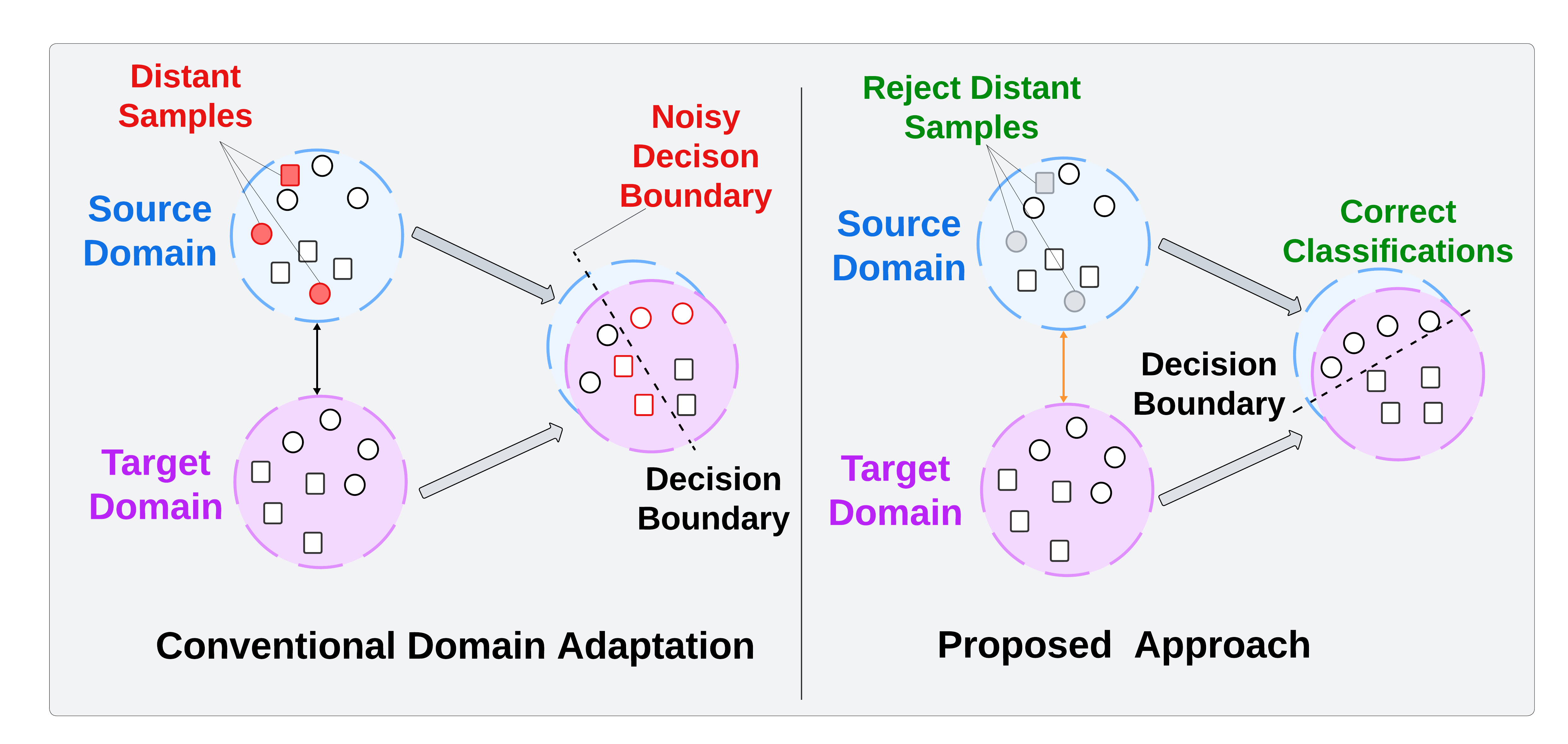}
    \caption{Left: Existence of distant samples causes conventional domain adaptation to learn a noisy decision boundary. Right: Correct classifications are attained by excluding distant samples from the adaptation.}
    \label{fig:intro2}
\end{figure}

As illustrated in Fig. \ref{fig:intro2}, in distant domain adaptation problems, there exist distant samples in the source domain, which causes the data distribution to be significantly different from the target domain. These samples negatively impact the knowledge transfer when learning the target domain decision boundary, causing conventional DA methods to fail. It is essential to identify the distant samples and reject them from the training to prevent the issue.

To address this problem, we propose an innovative Online Selective Adversarial Alignment (OSAA) method. Initially, our dynamic online strategy discerns less-relevant source domain samples, masking their loss gradients during training. This real-time calibration, free from offline pre-analysis or prior domain knowledge, ensures model focus on relevant samples. Furthermore, we introduce an intermediate domain, sampled dynamically from both source and target domains, to ease the adaptation across divergent domains. Lastly, to account for label disparities between domains, we develop a conditional adversarial adaptation technique, ensuring label consideration during the adaptation process.
To the best of our knowledge, this is the first work to define and investigate the distant domain adaptation problem specifically within fault diagnosis applications, with a particular focus on the challenges posed by negative transfer. The primary contributions of our research are:

\begin{itemize}
\item Formulation of the `distant domain adaptation' problem in intelligent fault diagnosis, with a comprehensive exploration of the challenges in identifying and addressing negative transfer.
\item Introduction of the Online Selective Adversarial Alignment (OSAA) algorithm, designed to counteract negative transfer arising from significant domain shifts.
\item Empirical validation of our proposed OSAA's efficacy through extensive experiments and ablation studies on two real-world datasets encompassing nine distinct domain adaptation scenarios.
\end{itemize}

\section{related work}  %
\label{sec:literature}
\subsection{Domain Adaptation}
With one or multiple labeled source domains sharing the same kind of task as the target domain, domain adaptation algorithms conduct knowledge transfer between the different data distributions. In unsupervised domain adaptation (UDA) settings, the target domain labels are completely not available. Discrepancy-based and adversarial-based methods form two main branches of the UDA algorithms\cite{b.18}. Discrepancy-based algorithms \cite{b.19, b.20, b.21} aim to minimize statistical distances between source and target domains. Adversarial-based algorithms \cite{b.22, b.23, b.24} conduct domain alignment with the help of domain discriminators and encourage the extraction of domain-invariant features. Besides, Adversarial Spectral Kernel Matching (AdvSKM)\cite{b.25} designs a hybrid spectral kernel network specific to time-series data to characterize non-stationary and non-monotonic statistics. Sparse Associative Structure Alignment (SASA) \cite{b.26} discovers causal structure in time-series data by sparse attention mechanisms.

\subsection{Domain Adaptation for Fault Diagnosis}
Bridging the theoretical foundations of domain adaptation to practical applications, particularly in machine fault diagnosis, is crucial for realizing its full potential. The majority of existing works are fault diagnosis scenarios under cross-working conditions. Lu \textit{et al.} \cite{b.15} first propose an autoencoder based on fully connected layers to minimize the domain discrepancy. Yu \textit{et al.} \cite{b.14-2} demonstrate the superior performance of 1D-CNN than typical fully connected layers for the feature extraction of gearbox fault signals. The integration of the self-attention mechanism \cite{b.16} into the feature extraction network has proved successful. Subdomain alignment \cite{b.8, b.17}, utilizing pseudo labels on the target domain, is effective in the cross-domain fault diagnosis field. Adversarial-based methods \cite{b.9, b.17-2, b.neww} implicitly reduce domain discrepancy by extracting indistinguishable and robust features for the domain classifier. While cross-working condition data contain machine-biased features, cross-machine fault analysis is more challenging, aiming to extract shared fault-related features. Li \textit{et al.} \cite{b.26.9} design a new autoencoder structure for the cross-machine analysis with different fault severities on multiple locations.
Zhu \textit{et al.} \cite{b.10} leverage a multi-adversarial learning strategy on multiple source machines for the target machine analysis. However, these studies are still focusing on the domains with mild discrepancy. There is limited relevant research addressing the potential issue of negative transfer under severe domain discrepancy, such as our raised cross-damage type fault diagnosis.

\subsection{Distant Transfer Learning}
There is limited research on distant transfer learning problems. When the domain gap is substantial, negative transfer occurs due to the mismatch in feature representations and variation in task complexity. Tan \textit{et al.} \cite{b.27} propose a selective learning algorithm for distant transfer learning with the help of a supervised autoencoder and intermediate domains. This approach is applied to various fields \cite{b.29, b.30}, such as image processing and text classification, when there are a few labeled target domain samples available. The limitation of these methods is to introduce the intermediate domain as data augmentation in a static way. Their requirement for some target domain labels does not apply to unsupervised settings. To the best of our knowledge, we are the first to address the fully unsupervised distant domain adaptation for fault diagnosis. 

\section{Methodology} \label{sec:me}
\begin{figure*}[h]
  \centering
  \includegraphics[width=\textwidth]{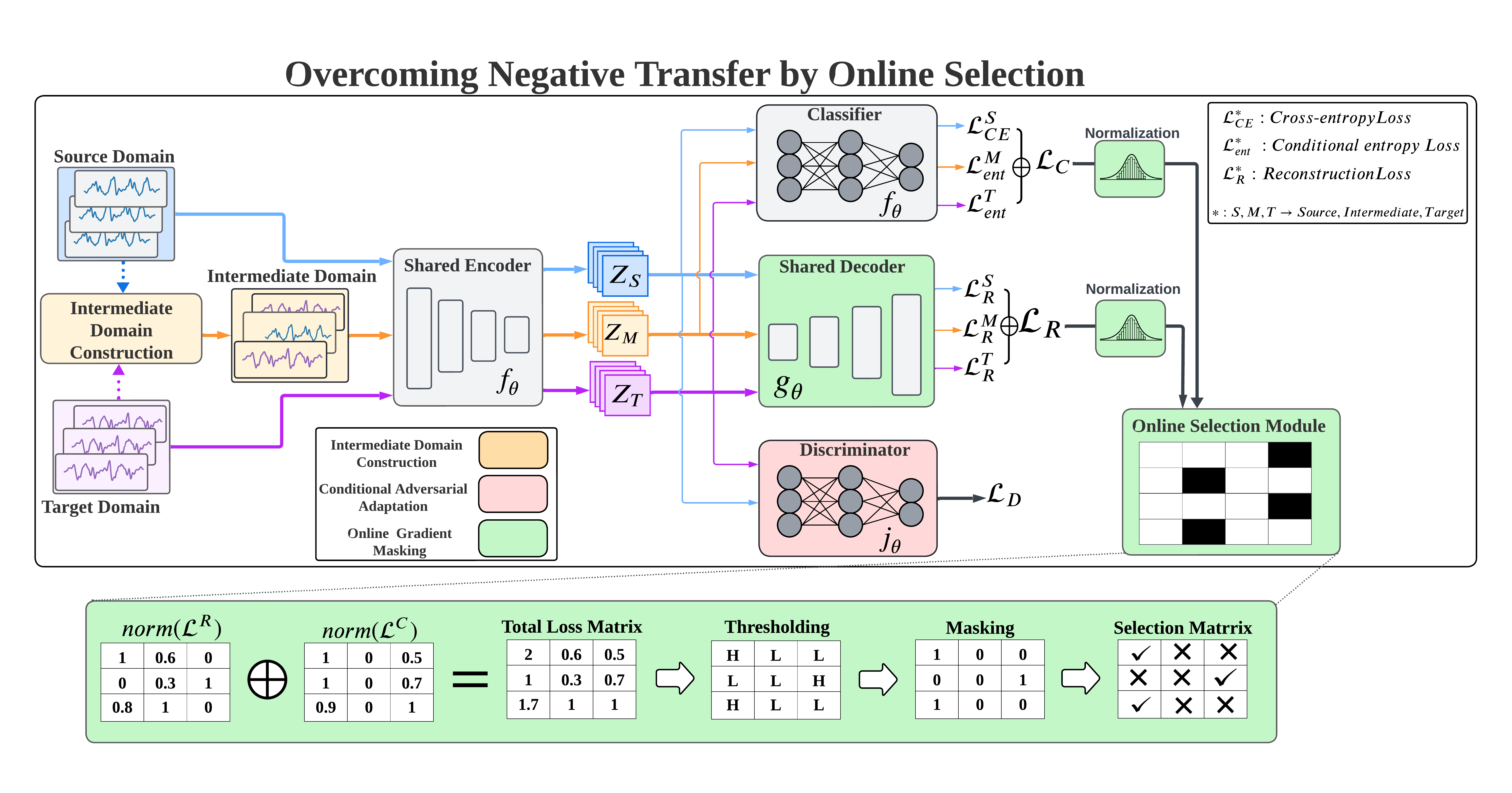}
\caption{
\label{fig:model}
The overview of our OSAA method to overcome negative transfer. Top: We illustrate the pipeline of the online selection mechanism. The source domain and the intermediate domain training samples are under selection. Bottom: We identify distant samples by observing training loss magnitudes (normalized) and conduct gradient selection by thresholding for the online gradient masking procedure.}
\end{figure*}
\subsection{Problem Formulation and Notations}

Let $\mathcal{D_S} = \{(\bm{x}^i_S, y^i_S)\}_{i=1}^{N_S}$ be a set of labeled source samples, drawn from the source distribution $P_S$, where $\bm{x}^i_S \in \mathcal{X}$ represents the input data and $y^i_S \in \mathcal{Y}$ is the corresponding label. Similarly, let $\mathcal{D_T} = \{(\bm{x}^j_T)\}_{j=1}^{N_T}$ represent a set of unlabeled target samples drawn from the target distribution $P_T$. $N_S$ and $N_T$ are the numbers of samples of source and target domains respectively. Here, we assume a distinct difference between the marginal distributions of the source and target domains, i.e., $P_S(X) \neq P_T(X)$, while the conditional distributions remain stable, that is, $P_S(Y|X) = P_T(Y|X)$. The aim of unsupervised domain adaptation (UDA) is to harness the knowledge acquired from the labeled source domain $\mathcal{D_S}$ to enhance the predictive performance on the unlabeled and shifted target domain $\mathcal{D_T}$. However, this research tackles a more intricate variant of UDA characterized by a substantial distribution shift between the source and target domains, introducing a critical issue known as `negative transfer'. Negative transfer arises when applying knowledge from the source domain leads to performance degradation in the target domain.
\subsection{Overview}
This section introduces the overview of the Online Selective Adversarial Alignment (OSAA) approach, illustrated in Figure \ref{fig:model}, specifically tailored to address the challenges of distant domain adaptation in fault diagnosis tasks. The approach comprises three main components: (1) an Online Selection Module responsible for identifying and excluding distant samples during the adaptation process; (2) an Intermediate Domain Construction Module serving as a transitional domain between the source and target; and (3) a Conditional Adversarial Alignment Module, which integrates label information during the adversarial adaptation process.
Subsequent subsections delve into a comprehensive explanation of each component.

\subsection{Online Selection Module}
Distant source samples can significantly contribute to the negative transfer phenomenon, thereby undermining adaptation performance. To mitigate this, we introduce an online selection mechanism designed to discern and disregard these distant samples, ensuring a focused adaptation using only relevant samples across domains. To do so, we propose a novel procedure that includes intermediate domain construction, input reconstruction task, feature classification task, and gradient masking to identify and isolate distant samples. Next, we will provide details for each step.

\subsubsection{Intermediate Domain Construction}
To bridge the inherent gap between the source and target domains, we introduce an intermediate domain that is specifically crafted to capture shared features, thereby mitigating the domain shift. This strategically constructed domain serves as a conduit, seamlessly connecting the source and target by minimizing their inherent disparities. Formally, given the source domain \( \mathcal{D_S} \)  with size \( N_S \) and the target domain \( \mathcal{D_T} \) with size \( N_T \), our aim is to construct the intermediate domain \( \mathcal{D_M} \). We achieve this by sampling 50\% of instances from each of the source and target domains. Specifically, we select a subset \( \mathcal{D}_{S}^{sub} \) of size \( 0.5 \times N_S \) from \( \mathcal{D_S} \) and another subset \( \mathcal{D}_{T}^{sub} \) of size \( 0.5 \times N_T \) from \( \mathcal{D_T} \). Thus, the intermediate domain \( \mathcal{D_M} \) is constructed as the union of these subsets, yielding \( \mathcal{D_M} = \mathcal{D}_{S}^{sub} \cup \mathcal{D}_{T}^{sub} \), resulting in an overall size \( N_M = 0.5 \times N_S + 0.5 \times N_T \).  

\subsubsection{Input Reconstruction Task}
Given raw machine fault signals $\bm{x}\in \mathbb{R}^m$, the encoder network $f_{\theta}$ compresses these signals into hidden representations $\bm{h}\in \mathbb{R}^h$. The decoder network $g_{\theta}$ then recovers the original signal from the extracted features through transposed convolutional layers and upsampling operations as follows: 
\begin{equation}\label{eq:h}
 \bm{h}_*  = f_{\theta}(\bm{x}_*),
\end{equation}
\begin{equation}\label{eq:reconsx}
   \hat{\bm{x}}_* = g_{\theta}(\bm{h}_*).
\end{equation}
Here, the subscript * indicates either the source, intermediate, or target domain. $\hat{\bm{x}}_*$ indicates the reconstructed features obtained by the decoder network. The encoder-decoder framework is updated by the reconstruction loss, which measures the dissimilarity between the pair of the raw signal and the reconstructed signal, revealing the quality of the feature extraction and reconstruction. The encoder-decoder parameters are shared for all three domains for consistency across domains. The final reconstruction loss is formulated as follows for all three domains:
\begin{equation}
\mathcal{L}_{R}=\sum\limits_{i=1}^{n} \left\|\hat{\bm{x}}^i_*-\bm{x}^i_*\right\|_2^2 .
\label{eq1}
\end{equation}

\subsubsection{Feature Classification Task}
Given the latent feature representations for the source domain (\( \bm{h}_S \)), the target domain (\( \bm{h}_T \)), and the intermediate domain (\( \bm{h}_M \)), our primary objective is to learn the fault diagnosis task. To achieve this, we initially exploit the labeled data available in the source domain to train both the encoder network \( f_\theta \) and the classifier network \( h_\theta \) using the cross-entropy loss.

\begin{equation}\label{eqyhat}
   \hat{y}_* = h_{\theta}(\bm{h}_*).
\end{equation}

Formally, for a true label \( y \) and its corresponding predicted probability \( \hat{y} \), the cross-entropy loss can be expressed as:

\begin{equation}
\mathcal{L}_{CE}(y, \hat{y}) = -\sum_{i}^{C} y_i \log(\hat{y}_i),
\label{eq:cross_entropy}
\end{equation}
where the summation spans all the $C$ classes in the dataset.

In the absence of ground truth labels for the intermediate and target domains, we utilize conditional entropy loss to encourage the classifier to produce confident predictions across both domains.

\begin{equation}
\mathcal{L}_{ent}(\mathbf{p}) = -\sum_{i=1}^{C} p_i \log(p_i),
\end{equation}
where \( p_i \) represents the softmax probabilities, derived as $p_i = \frac{\exp(\bm{h}_i)}{\sum_{j=1}^{C} \exp(\bm{h}_j)}$, with \( \bm{h}_i \) denoting the classifier's raw output (logit) for class \( i \).

The total classification loss is formulated as follows:
\begin{equation}
\mathcal{L}_{C} = \sum\limits_{i=1}^{n}  \mathcal{L}_{CE}(\hat{y}_S, y_S) + \mathcal{L}_{ent}(\hat{\bm{p}}_M) + \mathcal{L}_{ent}(\hat{\bm{p}}_T),
\label{eq2}
\end{equation}
where $\hat{\bm{p}}_M$ and $ \hat{\bm{p}}_T$ represent the predicted probabilities for the intermediate domain and target domain, respectively. 

\subsubsection{Online Selective Gradient Masking}
\begin{algorithm}[h]
\DontPrintSemicolon
\KwIn{Reconstruction Loss $\mathcal{L}_R$, Classification Loss $\mathcal{L}_C$}
\KwOut{Selected gradients for back-propagation}

\For{batch in Batches}{
    Min-max normalization of $\mathcal{L}_R$ and $\mathcal{L}_C$\;
    Summation of normalized $\mathcal{L}_R$ and $\mathcal{L}_C$\;
    Obtain masking matrices $V_S$ and $V_S$ via Eq. \ref{1}\;
    Dot Product of masking matrices $V_S$ and $V_M$ with gradient matrices \(\nabla_S\) and \(\nabla_M\)\;
    Get the masked gradient matrices $\hat{\nabla}_S$ and $\hat{\nabla}_M$
}
\Return {$\hat{\nabla}_S$ , $\hat{\nabla}_M$}
\caption{Gradient Selection}
\label{algo2}
\end{algorithm}

 As established in prior works \cite{b.27, b.31}, samples with higher training loss are more likely to be unreliable and, thus, less beneficial for the final task. In the context of distant domain adaptation, achieving a stronger performance in the source domain does not guarantee a better result in the target domain. To counteract this, during training, we utilize the total classification loss \(\mathcal{L}_C\) and the total reconstruction loss \(\mathcal{L}_R\) to construct binary masking matrices \(V_S\) for the source domain and \(V_M\) for the intermediate domain. We present the computation of the masking matrices as follows: 
 \begin{equation}
V_S, V_M = 
\begin{cases}
0 & \text{if } \lVert \mathcal{L}_R \rVert + \lVert \mathcal{L}_C \rVert \geq Q_{p}(\lVert \mathcal{L}_R \rVert + \lVert \mathcal{L}_C \rVert), \\
1 & \text{otherwise},
\end{cases}
\label{1}
\end{equation}

where the threshold $Q_p$ represents the $p$-th percentile of the distribution of normalized loss summation value across the training batch. 

Subsequently, to suppress the gradients of distant samples, we perform a dot product between the masking matrices and the gradient matrices \(\nabla_S\) and \(\nabla_M\) corresponding to the source and intermediate domains, respectively. The gradient matrices are calculated by the differentiation of loss functions $\mathcal{L}_R$ and $\mathcal{L}_C$ with respect to network parameters $f_{\theta}$, $ g_{\theta}$ and $h_{\theta}$. The suppressed gradients are given by \(\hat{\nabla}_S = V_S \cdot \nabla_S\) and \(\hat{\nabla}_M = V_M \cdot \nabla_M\). During each training step, we first fix all network parameters and then update the gradient masking matrices \(V_S\) and \(M_M\). After this, only the selected gradients  \(\hat{\nabla}_S\) and \(\hat{\nabla}_M\) are back-propagated to refine the network parameters. This process ensures that the model consistently emphasizes relevant samples beneficial for the target domain learning.
 
\begin{algorithm}[h]
\DontPrintSemicolon
\KwIn{Source domain $D_S$, Intermediate domain $D_M$, Target domain $D_T$\\
Initialized encoder $f_{\theta}$, decoder $g_{\theta}$, task classifier $h_{\theta}$, domain discriminator $j_{\theta}$}
\KwOut{Trained encoder $f_{\theta}$ and task classifier $h_{\theta}$}

\For{number of epochs}{
    Sample mini-batch of samples from all domains $X_S \sim P_S$, $X_M \sim P_M$, $X_T \sim P_T$\;
    Pass the sampled batch to $f_{\theta}$ for feature extractions $\bm{h}_S, \bm{h}_M, \bm{h}_T$ by Eq. \ref{eqyhat}\;
    Pass the feature extractions to $h_{\theta}$ for label predictions $\hat{y}_S, \hat{y}_M, \hat{y}_T$ by Eq. \ref{eq:reconsx}\;
    Pass the feature extractions to $j_{\theta}$ for label-conditioned domain predictions $\hat{d}_s, \hat{d}_t$ by Eq. \ref{eq666}\;
    Pass the feature extractions to $g_{\theta}$ for feature reconstructions $\hat{X}_S, \hat{X}_M, \hat{X}_T$ by Eq. \ref{eq:reconsx}\;
    Compute domain discrimination loss $\mathcal{L}^D$ by Eq. \ref{eq:LD}\;
    Update selection variables $v_S, v_M$ by batch-level normalization and portion hyperparameter $p$\;
    Compute reconstruction loss $\mathcal{L}_R$ by Eq. \ref{eq1}\;
    Compute task classification loss $\mathcal{L}_C$ by Eq. \ref{eq2}\;
    Conduct gradient selection as in Algorithm \ref{algo2}\;
    Update by the overall objective function by Eq. \ref{eq_total}\;
}
\Return{$f_{\theta}$, $h_{\theta}$}

\caption{Online Selective Adversarial Alignment}
\label{algo1}
\end{algorithm}
\subsection{Adversarial Alignment Module}
Our domain adaptation strategy employs an adversarial approach to narrow the domain disparity between source and target samples. Traditional adversarial methods often align the feature distributions of source and target domains without integrating the task-related label information. To harness this label information, we propose the Conditional Adversarial Alignment, inspired by \cite{b.23}. Specifically, we feed both the features \(\bm{h}\) and their corresponding predicted labels \(\hat{y}\) into the discriminator network \(j_{\theta}\). This inclusion allows the discriminator to factor in task-related information during the alignment phase, representing this as:
\begin{equation}\label{eq666}
   \hat{d}_* = j_{\theta}((\bm{h}_*, \hat{y}_*)).
\end{equation}

In this adversarial framework, the encoder \(f_{\theta}\) and the domain discriminator \(j_{\theta}\) are jointly trained. The discriminator aims to discern the origin (source or target) of a given data sample, considering its predicted label, while the encoder's objective is to produce features that are domain-agnostic, complicating the discriminator's task. The loss for the conditional discriminator is:
\begin{equation} \label{eq:LD}
\mathcal{L}_{D} = \mathbb{E}_{\bm{x_S} \sim \mathcal{D}_S} \log(\hat{d}_S) + \mathbb{E}_{\bm{x_T} \sim \mathcal{D}_T} \log(1 - \hat{d}_T).
\end{equation}

During training, while the discriminator minimizes \(\mathcal{L}_{D}\), the encoder attempts to maximize it, effectively training with inverse labels. By integrating label predictions with the adversarial alignment, our Conditional Adversarial Alignment not only fosters domain-invariant feature extraction but also exploits task-related label information, yielding a more effective domain adaptation strategy.

\subsection{Overall Objective Function}
In the final step, we integrate all the individual networks, encoder $\bm{f}_{{\theta}}$, decoder $\bm{g}_{{\theta}}$, task classifier $\bm{h}_{{\theta}}$, and domain discriminator $j_{\theta}$, to formulate the overall objective function with weighted sum operation: 

\begin{equation}
\begin{aligned}
& \mathop{\min}_{{f}_{{\theta}}, {g}_{{\theta}}, {h}_{{\theta}}}  \widehat{\mathcal{L}}_{R} + \lambda_1 \cdot  \widehat{\mathcal{L}}_{C} - \lambda_2 \cdot \mathcal{L}_{D} \\
& \mathop{\min}_{j_{\theta}} \mathcal{L}_{D},
\label{eq_total}
\end{aligned}
\end{equation}
where $\lambda_0, \lambda_1, \lambda_2 \in [0,1]$ are the loss weight hyperparameters. $\widehat{\mathcal{L}}_{R}, \widehat{\mathcal{L}}_{C}$ are the losses after the gradients of $\mathcal{L}_{R}, \mathcal{L}_{C}$ being selected. For the evaluation phase, the test samples are passed through the encoder $\bm{f}_{{\theta}}$ for feature extraction, and then fed into the task classifier $\bm{h}_{{\theta}}$ for the final label predictions as follows:
\begin{equation}
   \hat{y}_{test} = h_{\theta}(f_{\theta}(\bm{x}_{test})).\label{eqtest}
\end{equation}
To encapsulate the steps involved in our methodology, we present the entire pipeline in Algorithm \ref{algo1}, providing a step-by-step guide and the logical flow of our proposed model.

\section{Experimental Setup}
\label{sec:exp}
In this section, we present a comprehensive evaluation of our OSAA method across three distinct unsupervised domain adaptation scenarios. Through extensive experimentation, we provide both quantitative and qualitative results to demonstrate the effectiveness of our approach.
\subsection{Datasets}
\subsubsection{Paderborn University dataset}
The Paderborn University dataset \cite{b2} is a comprehensive collection of machine fault data. It captures variations in three key parameters: damage type, rotation speed, and loading torque. These parameters, in their various combinations, comprise the six distinct working conditions represented in the dataset, with their details shown in Table \ref{table1}. Specifically, the damage type includes both artificial damages, intentionally induced by certain devices, and real damage faults resulting from accelerated run-to-failure tests. Each domain within the dataset has a consistent sample length of 5,120. 
\begin{table}[h]
\centering
\caption{Details of the Paderborn University (PU) Dataset}
\setlength{\tabcolsep}{3pt}
\label{table1}
    \begin{tabular}{c|ccc}
     \toprule
   Domain & Damage Type & Loading Torque (nm) & Radial Force (N) \\
   \midrule
   A & Artificial & 0.1 &1000\\ 
   B & Artificial & 0.7 &400\\ 
   C & Artificial & 0.7&1000\\ 
   D & Real & 0.1&1000\\ 
   E & Real & 0.7&400\\ 
   F & Real & 0.7&1000\\ 
   \bottomrule
    \end{tabular}
\end{table}

\subsubsection{ Case Western Reserve University (CWRU) dataset}
The Case Western Reserve University (CWRU) dataset \cite{b3} is a well-known bearing machinery dataset out of real damages. The test rig for the dataset has accelerometers placed at both the drive end and the fan end. The subset domains picked for the domain adaptation experiment in Table \ref{tablecwru} are collected with 12k samples per second. Data samples have a fixed length of 1024. The CWRU dataset encompasses machine fault data under various working conditions across different motor speeds and fault diameters.

\begin{table}[htbp]
\centering
\caption{Details of the Case Western Reserve University (CWRU) Dataset}
\setlength{\tabcolsep}{3pt}
\label{tablecwru}
    \centering
    \begin{tabular}{c|cccc}
     \toprule
   Domain & Location & Motor Speed (rpm) & Diameter (inches) \\
   \midrule
   FE1730 (F30)& Fan End & 1730 & 0.007, 0.021\\ 
   FE1772 (F72)& Fan End & 1772 & 0.007, 0.021\\ 
   FE1797 (F97)& Fan End & 1797 & 0.007, 0.021\\ 
   DE1750 (D50)& Drive End & 1750& 0.007, 0.021\\ 
   DE1772 (D72)& Drive End &  1772& 0.007, 0.021\\ 
   DE1797 (D97)& Drive End & 1797& 0.007, 0.021\\ 
   \bottomrule
    \end{tabular}
\end{table}

\subsection{Cross-domain Settings}
Three domain adaptation settings have been formulated, as shown in Table \ref{table:setting}. The Paderborn dataset informs the first two settings. The initial setting designates artificial damage domains as the source and real damage domains as the target, a configuration referred to as "artificial-to-real." In contrast, the subsequent setting adopts the "real-to-artificial" configuration. The third setting, derived from the CWRU dataset, addresses variations in motor speed and the specific measurement locations, whether at the fan-end or the drive-end.

\begin{table}[h]
\centering
\caption{Experimental Groups and Scenarios}
\setlength{\tabcolsep}{6pt}
\label{table:setting}
    \begin{tabular}{ccc}
     \toprule
     Paderborn&Paderborn&CWRU \\
   \midrule
  Artificial-to-real &Real-to-artificial&Cross-speed\&location \\
 B $\rightarrow$ D & D $\rightarrow$ B & F72 $\rightarrow$ D97\\ 
   A $\rightarrow$ E & E $\rightarrow$ A &  F30 $\rightarrow$ D50\\
    C $\rightarrow$ F & F $\rightarrow$ C &  D72 $\rightarrow$ F97   \\
    \bottomrule
    \end{tabular}
\end{table}

\subsection{Implementation Details}
We implement an encoder model with three 1D-CNN layers, each having 64 channels, a kernel size of 9, and a stride of 1. The dropout probability was set to 0.4. Following this, an adaptive average pooling layer adjusted the hidden representation to a dimension of 128, as described in ADATIME \cite{b.18}.  The decoder includes three transposed 1D-convolutional layers and three max un-pooling layers for upsampling. The task classifier contains a single hidden layer that maps the hidden representation to the label space. Additionally, the domain discriminator has a hidden layer with a size of 128. The model parameters were set with a learning rate of \(1 \times 10^{-4}\), weight decay of \(1 \times 10^{-5}\), and a batch size of 64 for both encoders and decoders. The selection proportion in the Online Selection Module for both source and intermediate domains was 50\%. Each training process was conducted over 20 epochs, with all loss weights set to 1. It's worth noting that we used similar settings for baseline methods to ensure a consistent comparison. Each experiment was repeated five times with different random seeds.
\section{Results and Discussions}
\subsection{Baselines}
 As introduced in Section \ref{sec:literature}, our method is compared with a wide variety of other unsupervised domain adaptation algorithms, including discrepancy-based methods DDC \cite{b.19}  and DSAN \cite{b.20}, adversarial-based methods DANN \cite{b.22}, CDAN \cite{b.23} and DIRT-T \cite{b.24}, time series-specific methods AdvSKM \cite{b.25} and SASA \cite{b.26}. We also include a source-only model as the non-transfer baseline for comparison. The source-only model consisting of a feature extractor and a task classifier is trained on the source domain data only. 

\subsection{Comparison with baselines}
We compare the performances of our proposed OSAA with related state-of-the-art domain adaptation algorithms, as shown in Table \ref{table: mainres1},\ref{table: mainres2}, and \ref{table: mainres3}.
\begin{table}[hbt!]
\centering
\caption{Results of PU (Artificial-to-Real) Scenarios}
\setlength{\tabcolsep}{5pt}
\label{table: mainres1}
    \begin{tabular}{c|ccc|c}
    \toprule
    Methods & B $\rightarrow$ D & A $\rightarrow$ E & C $\rightarrow$ F & AVG \\
    \midrule
    SourceOnly & 46.02$\pm$2.02 & 37.84$\pm$1.07 & 35.09$\pm$1.92 & 39.65 \\
    \midrule
    DIRT-T & 23.70$\pm$2.57 & 20.96$\pm$0.79 & 17.05$\pm$1.93 & 20.57 \\
    DSAN & 26.82$\pm$1.83 & 36.10$\pm$2.97 & 39.51$\pm$2.65 & 34.14 \\
    SASA & 28.86$\pm$1.16 & \underline{45.79$\pm$2.54} & 39.46$\pm$3.22 & 38.04 \\
    DDC & 47.20$\pm$3.61 & 30.04$\pm$1.78 & 44.84$\pm$2.84 & 40.70 \\
    AdvSKM & \underline{53.00$\pm$4.01} & 31.82$\pm$1.31 & \underline{50.53$\pm$2.64} & \underline{45.12} \\
    DANN & 24.95$\pm$3.24 & 25.18$\pm$2.64 & 47.86$\pm$2.47 & 32.60 \\
    CDAN & 36.19$\pm$2.13 & 27.74$\pm$2.08 & 39.07$\pm$1.99 & 34.33 \\
    \midrule

    \textbf{OSAA} & \textbf{64.66$\pm$1.31} & \textbf{52.35$\pm$1.11} & \textbf{65.82$\pm$1.62} & \textbf{60.94} \\
    \bottomrule
    \end{tabular}
\end{table}

\begin{table}[hbt!]
    \centering
    \caption{Results of PU (Real-to-Artificial) Scenarios}
\setlength{\tabcolsep}{5pt}
\label{table: mainres2}
    \begin{tabular}{c|ccc|c}
    \toprule
    Methods & D$\rightarrow$B & E $\rightarrow$ A & F $\rightarrow$ C & AVG \\
    \midrule
    SourceOnly & 16.01$\pm$2.05 & 21.86$\pm$2.18 & 17.74$\pm$3.82 & 18.54 \\
    \midrule
    DIRT-T & 25.61$\pm$0.79 & 24.99$\pm$0.16 & \underline{25.92$\pm$0.69} & 25.51 \\
    DSAN & 22.93$\pm$2.38 & 15.67$\pm$1.33 & 17.48$\pm$2.32 & 18.68 \\
    SASA & 18.98$\pm$3.22 & 17.73$\pm$0.76 & 25.46$\pm$0.28 & 20.72 \\
    DDC & 17.75$\pm$1.19 & 26.06$\pm$0.86 & 24.82$\pm$0.23 & 22.88 \\
    AdvSKM & 17.06$\pm$2.52 & 17.09$\pm$2.18 & 23.55$\pm$2.07 & 19.23 \\
    DANN & 14.31$\pm$1.65 & 17.89$\pm$2.25 & 18.63$\pm$1.79 & 16.94 \\
    CDAN & \underline{28.81$\pm$1.46} & \underline{27.55$\pm$1.26} & 24.73$\pm$1.58 & \underline{27.03} \\
    \midrule
    \textbf{OSAA} & \textbf{39.09$\pm$2.37} & \textbf{29.99$\pm$2.64} & \textbf{35.78$\pm$1.95} & \textbf{34.95} \\
    \bottomrule
    \end{tabular}
\end{table}

\begin{table}[hbt!]
    \centering
\caption{Results of CWRU Scenarios}
\setlength{\tabcolsep}{5pt}
\label{table: mainres3}
    \begin{tabular}{c|ccc|c}
    \toprule
    Methods & F72 $\rightarrow$ D97 & F30 $\rightarrow$ D50 & D72 $\rightarrow$ F97 & AVG \\
    \midrule
    SourceOnly & 35.14$\pm$1.44 & 28.55$\pm$1.86 & 32.58$\pm$1.32 & 32.09 \\
    \midrule
    DIRT-T & 45.39$\pm$0.98 & 27.93$\pm$2.02 & \textbf{48.01$\pm$2.23} & 40.45 \\
    DSAN & 40.10$\pm$1.94 & 28.49$\pm$1.66 & 46.64$\pm$2.03 & 38.41 \\
    SASA & 37.68$\pm$2.10 & 35.09$\pm$1.63 & 39.78$\pm$1.78 & 37.52 \\
    DDC & 44.08$\pm$2.09 & 18.67$\pm$3.55 & 38.42$\pm$1.66 & 33.72 \\
    AdvSKM & 32.30$\pm$3.19 & 28.91$\pm$2.86 & 23.38$\pm$3.11 & 28.20 \\
    DANN & 42.09$\pm$1.59 & \underline{41.50$\pm$2.14} & 41.48$\pm$2.92 & 41.69 \\
    CDAN & \underline{49.22$\pm$2.03} & 35.21$\pm$2.11 & 45.94$\pm$1.83 & \underline{42.46} \\
    \midrule
    \textbf{OSAA} & \textbf{49.34$\pm$1.35} & \textbf{49.88$\pm$1.36} & \underline{46.96$\pm$2.62} & \textbf{48.73} \\
    \bottomrule
    \end{tabular}
\end{table}

\begin{figure*}
\centering
\begin{subfigure}[t]{0.5\columnwidth}
    \centering
    \includegraphics[width=\linewidth]{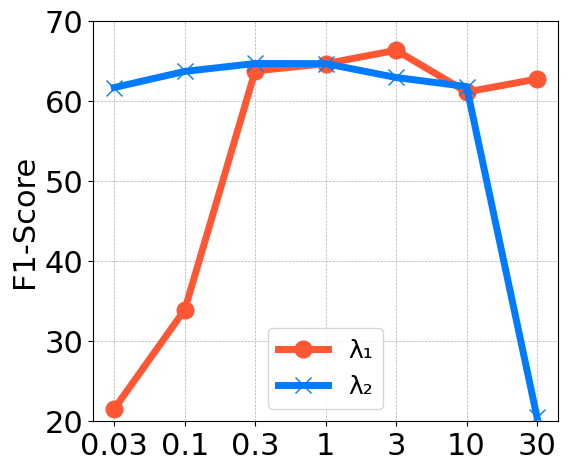}
    \caption{B$\rightarrow$D}
\end{subfigure}%
\begin{subfigure}[t]{0.5\columnwidth}
    \centering
    \includegraphics[width=\linewidth]{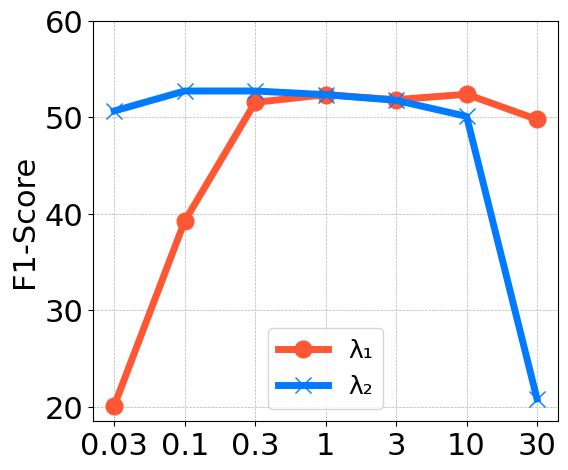}
    \caption{A$\rightarrow$E}
\end{subfigure}%
\begin{subfigure}[t]{0.5\columnwidth}
    \centering
    \includegraphics[width=\linewidth]{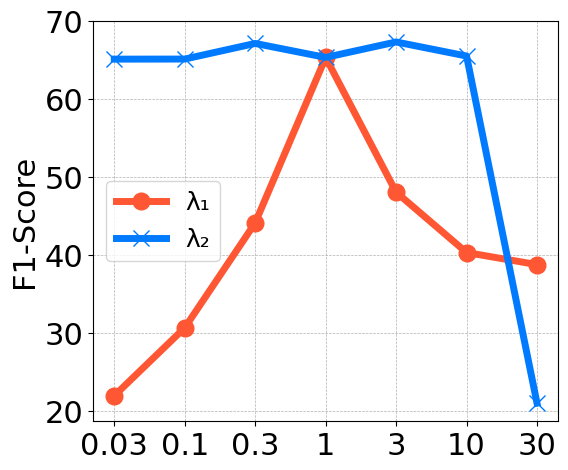}
    \caption{C$\rightarrow$F}
\end{subfigure}

\caption{F1-score against different combinations of loss weights, with $\lambda_1$ and $\lambda_2$ ranging from 0.03 to 30 respectively.}
\label{fig:loss}
\end{figure*}

Our OSAA method consistently surpasses other state-of-the-art domain adaptation algorithms across all three settings, achieving the highest F1-score in eight out of nine scenarios. Interestingly, in the Paderborn artificial-to-real scenarios, certain existing UDA algorithms register F1-scores even lower than the baseline set by the source-only model. This underperformance can be attributed to the negative transfer phenomenon, where distant source sample information adversely impacts the target domain. In contrast, OSAA adeptly filters out misleading information from the source domain, effectively countering negative transfer. Consequently, our model realizes significant performance enhancements, with improvements ranging from 7\% in the $A\rightarrow E$ scenario to 15\% in the $C\rightarrow F$ scenario.

\begin{table}[hbt!]
    \centering
    \caption{Ablation Studies of Key Components in PU (Artificial-to-Real) Scenarios}
\setlength{\tabcolsep}{5pt}
\label{table: ablation}
    \begin{tabular}{ccc|ccc|c}
    \toprule
    Selec & Intm & Disc&B $\rightarrow$ D&A $\rightarrow$ E&C $\rightarrow$ F&AVG\\
    \midrule
    $\bm{-}$ & \checkmark &\checkmark&36.19$\pm$2.14 & 27.74$\pm$2.09 & 39.07$\pm$1.99 & 34.33\\
    \checkmark &$\bm{-}$ & \checkmark &37.40$\pm$2.42&35.25$\pm$2.71&44.68$\pm$1.26&39.11\\
        \checkmark & \checkmark & $\bm{-}$ &\underline{58.23$\pm$1.27}&\underline{48.86$\pm$1.20}&\underline{60.36$\pm$1.63}&\underline{55.82}\\

        \checkmark & \checkmark &\checkmark&\textbf{64.66$\pm$1.31}&\textbf{52.35$\pm$1.11}&\textbf{65.82$\pm$1.62}&\textbf{60.94}\\
        \bottomrule
    \end{tabular}
\end{table}

\subsection{Ablation Study}
We conduct ablation studies to verify the efficacy of each key component of our algorithm under different configurations. We carry out experiments with the following control groups: the reduced model without domain discriminator (Disc), the reduced model without the selection mechanism (Selec), and the modified model without the constructed intermediate domain (Intm). Then we conduct a quantified analysis of the key feature of our algorithm: the selection portion. The results on the Paderborn dataset with the artificial-to-real setting are used as representatives.

\subsubsection{Effectiveness of Each Component}
In Table \ref{table: ablation}, we demonstrate the effectiveness of each component of the proposed OSAA model. Notably, the selection mechanism is pivotal, with its removal leading to a substantial performance drop of over 25\%. Nevertheless, the inclusion of the intermediate domain is equally crucial. Without the intermediate domain, the online selection underperforms the full OSAA model by over 20\%, underscoring the intermediate domain's role in enhancing performance and bridging the two distant domains. Lastly, excluding the domain discriminator results in a relatively minor performance decline of around 5\%. However, this result still emphasizes the importance of considering label distribution during domain alignment.

\begin{figure}[htbp]
  \centering
   \includegraphics[width=0.4\textwidth]{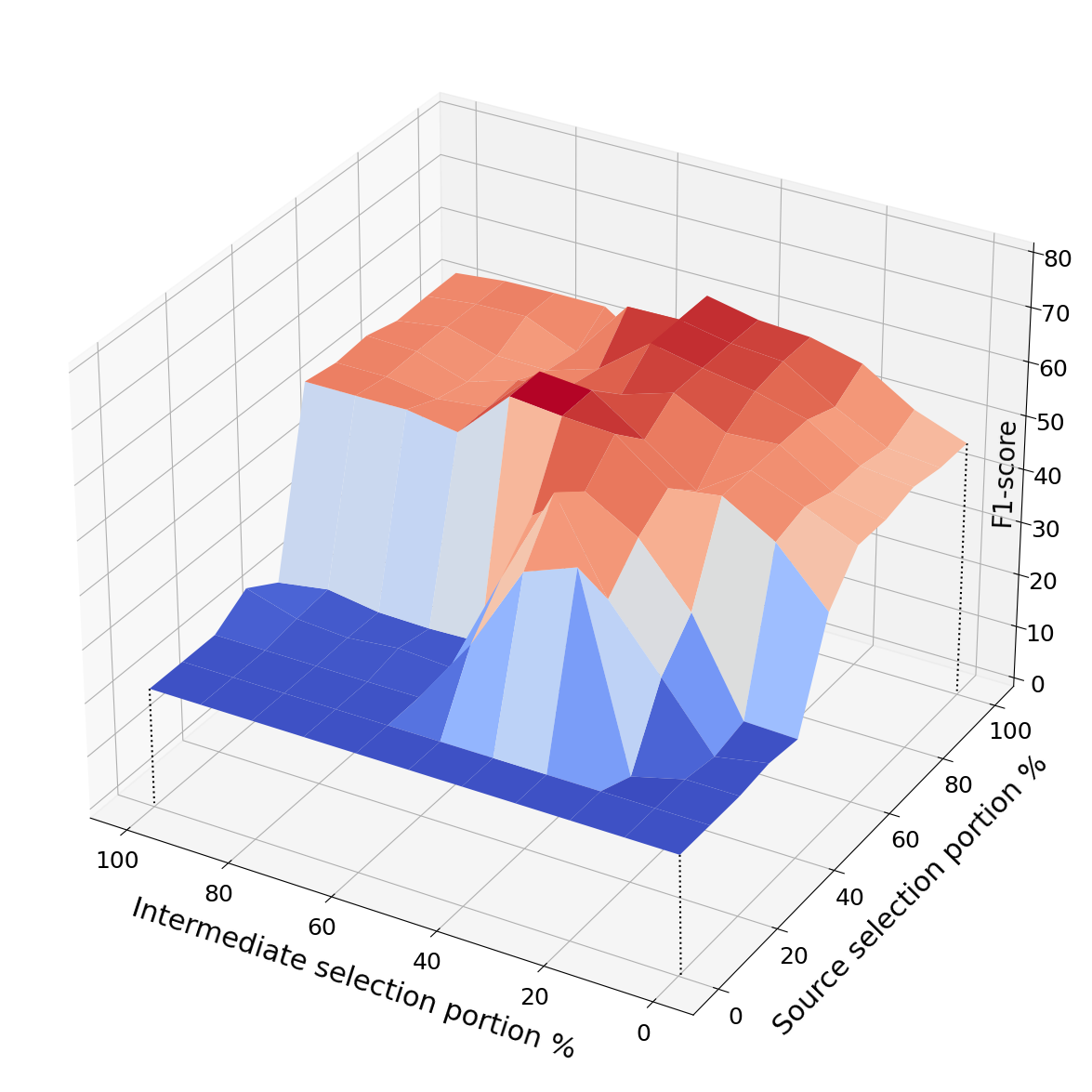}
   \caption{F1-score (vertical, bottom to top) versus selection portion ranging from 0 to 100$\%$ for the source (front axis, right to left) and intermediate (right axis, front to back) domain.}
   \label{fig:surface}
\end{figure}
\subsubsection{Appropriate Selection of Samples Portion}
We also conduct a detailed analysis of the selection portion hyperparameter $p$ on both the source domain and the intermediate domain, reinforcing the efficacy of the selective strategy. Fig. \ref{fig:surface} illustrates the impact of varying selection proportions of both domains on the model's performance. While the optimal score (dark red) is achieved by selecting $60\%$ samples from both source and intermediate domains, our reported performance is based on $50\%$ selection for the two domains, which is comparable to the optimal score. The overall results reveal a delicate balance of choosing $p$: an overly large $p$ leads to the inclusion of negative samples, resulting in a significant performance drop; an overly small $p$ leads to a more catastrophic performance due to lack of data. While determining the optimal selection proportion beforehand is challenging, our preliminary choice of 50\% appears to be sufficiently effective.

\subsection{Sensitivity Analysis}
We conducted a sensitivity analysis on the Paderborn University dataset to study the sensitivity of our approach to various loss weights $\lambda_1$ and $\lambda_2$, as defined in Eq. \ref{eq_total}. The effects of varying $\lambda_1$ for the classification loss $\mathcal{L}_C$ and $\lambda_2$ for the discriminator loss $\mathcal{L}_D$ are depicted in Fig. \ref{fig:loss}, with values spanning from 0.03 to 30. Initially, with $\lambda_1$ set to 1, the model consistently exhibited commendable performance, showing minimal sensitivity to alterations in the $\lambda_2$ value (represented in orange). Peak prediction accuracy was observed at $\lambda_2 = 0.3$. However, performance declined noticeably when $\lambda_2$ was increased to larger values, such as 30. In subsequent tests, with $\lambda_2$ held constant at $0.3$, optimal performance was observed around $\lambda_1 = 1$ (represented in blue). A decline in performance was noted when $\lambda_1$ was reduced below $0.3$, suggesting heightened sensitivity to lower $\lambda_1$ values. This trend highlights the significance of label-related supervision in the fault diagnosis task.

\section{Conclusion \label{sec:co}} 
In this study, we present a novel Online Selective Adversarial Alignment (OSAA) model tailored for distant domain adaptation challenges in fault diagnosis applications. To address the negative transfer issue, our proposed approach incorporates an Online Selection Module, responsible for selective gradient masking for every training batch based on the reconstruction loss and task classification loss, and an Adversarial Alignment Module, designed for domain-invariant feature extraction as well as implicit domain adaptation. Without the use of external data, we construct an intermediate domain as an augmentation to the selective training strategy. Using the Paderborn University and the Case Western Reserve University datasets, we carry out comprehensive experiments and ablation studies for the evaluation. We provide compelling evidence for the effectiveness and robustness of our proposed algorithm.

\end{document}